\documentclass[final,5p,times,twocolumn]{elsarticle}

\usepackage{amsmath}
\usepackage{amssymb}
\usepackage{orcidlink}
\usepackage{cleveref}
\usepackage{changes}
\usepackage{booktabs}
\usepackage{graphics}
\usepackage{tikz}
\usepackage{graphicx}
\usetikzlibrary{positioning}

\journal{}

\begin{document}

\begin{frontmatter}



\title{Direction-aware multi-scale gradient loss for infrared and visible image fusion}


\author[1,2,4]{Kaixuan Yang\orcidlink{0009-0002-5315-9894}}
\author[1,2]{Wei Xiang}
\author[1,2,4]{Zhenshuai Chen\orcidlink{0000-0002-3552-4002}}
\author[1,2,4]{Tong Jin\orcidlink{0009-0002-4568-8913}}
\author[1,2]{Yunpeng Liu\corref{cor1}}

\cortext[cor1]{This is the corresponding author. Email address: ypliu@sia.cn}

\affiliation[1]{organization={Key Laboratory of Opto-Electronic Information Processing, Chinese Academy of Sciences},
            addressline={Chuangxin Road}, 
            city={Shenyang},
            postcode={110016}, 
            state={Liaoning},
            country={China}}
            
\affiliation[2]{organization={Shenyang Institute of Automation, Chinese Academy of Sciences},
	addressline={Chuangxin Road}, 
	city={Shenyang},
	postcode={110016}, 
	state={Liaoning},
	country={China}}

\affiliation[4]{organization={University of Chinese Academy of Sciences, Chinese Academy of Sciences},
	addressline={Yanqi Lake East Road}, 
	city={Shenyang},
	postcode={100049}, 
	state={Liaoning},
	country={China}}

\begin{abstract}
Infrared and visible image fusion aims to integrate complementary information from co-registered source images to produce a single, informative result. Most learning-based approaches train with a combination of structural similarity loss, intensity reconstruction loss, and a gradient-magnitude term. However, collapsing gradients to their magnitude removes directional information, yielding ambiguous supervision and suboptimal edge fidelity. We introduce a direction-aware, multi-scale gradient loss that supervises horizontal and vertical components separately and preserves their sign across scales. This axis-wise, sign-preserving objective provides clear directional guidance at both fine and coarse resolutions, promoting sharper, better-aligned edges and richer texture preservation without changing model architectures or training protocols. Experiments on open-source model and multiple public benchmarks demonstrate effectiveness of our approach.
\end{abstract}




\end{frontmatter}


\section{Introduction}
\label{sec:introduction}

Multimodal image fusion can be regarded as a kind of image restoration, and infrared and visible image fusion is the most common one, which combines information from multiple sensors and multiple shooting configurations to generate a comprehensive representation of scenes and objects with more details, higher contrast, and improved visual effects~\cite{yang2024lfdt}. Infrared images provide thermal radiation-based information, more valuable for detecting thermal targets and observing nighttime activities; Visible images provide the reflectance-based visual information, akin to human vision but cannot present visual details under harsh conditions. The infrared and visible image fusion focuses on fusing the complementary information of infrared and visible images, yielding high-quality fusion images~\cite{yang2024review}. It is widely used in tasks such as scene information enhancement~\cite{yan2022learning,yan2022rignet,zhao2022discrete}, image registration~\cite{jiang2022towards,wang2022unsupervised,xu2022rfnet}, and downstream tasks such as object detection~\cite{liu2022target} and semantic segmentation~\cite{tang2022image} in scenes with multiple sensors.

\begin{figure*}[ht]
	\centering
	\includegraphics[width=1\textwidth]{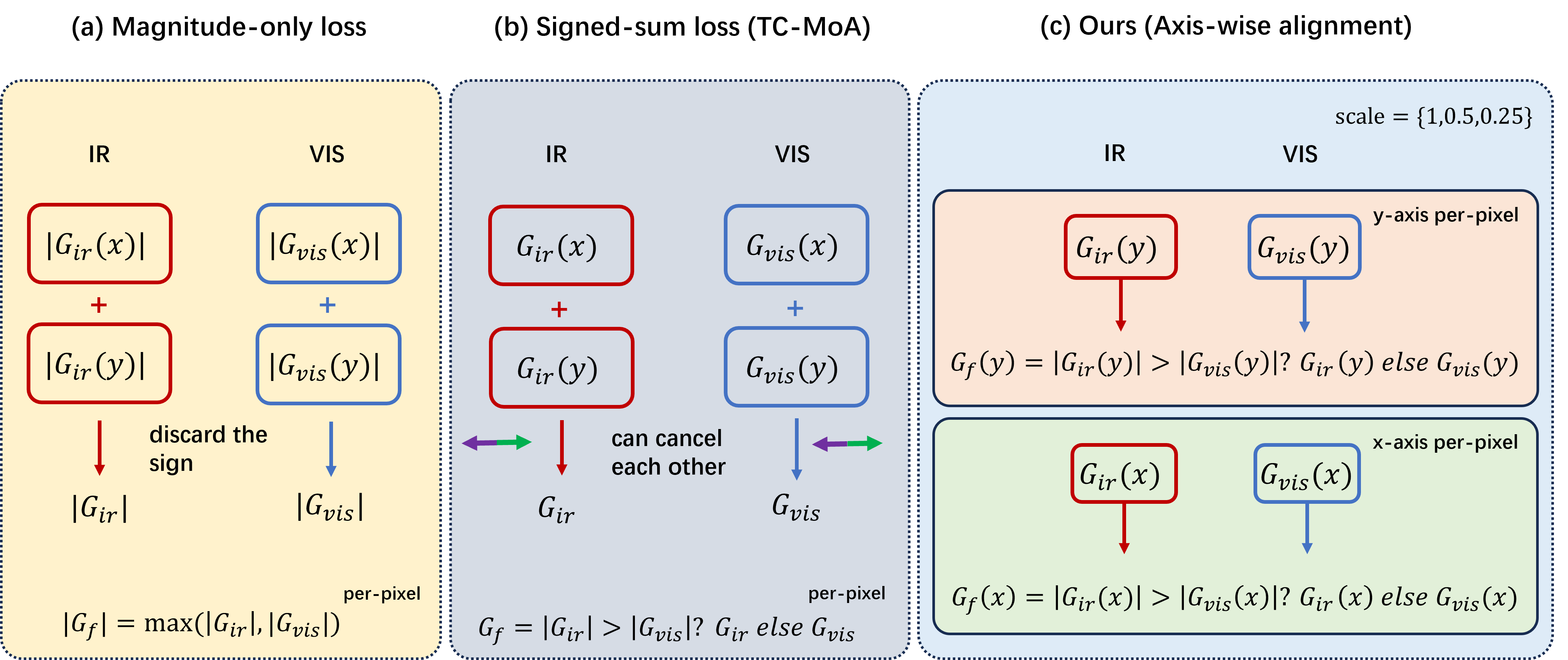}
	\caption{Comparison of gradient loss formulations. The proposed direction-aligned loss (c) performs independent supervision along horizontal and vertical directions, avoiding gradient cancellation and preserving sharper edge structures.}
	\label{fig:gradloss_compare}
\end{figure*}

Infrared and visible image fusion is an unsupervised task that relies on carefully designed loss functions to ensure that the fused output simultaneously preserves thermal targets from infrared images and fine spatial details from visible images. Because ground-truth fused images are generally unavailable, the training objective is typically formulated as a combination of multiple loss terms that jointly enforce pixel-level fidelity, structural similarity, and feature-level information retention from both modalities~\cite{zhangRethinkingImageFusion2020}.

In existing infrared and visible image fusion methods, two types of losses are most commonly used. The first is the intensity reconstruction loss, which computes the pixel-wise maximum of the infrared and visible image intensities and then measures the norm between the fused image’s intensity and this maximum map. The second is the gradient loss, where the pixel-wise maximum of the gradient magnitudes from the two source images is first computed, and the difference between this maximum gradient and that of the fused image is then minimized\cite{zhao2023cddfuse}. However, this conventional formulation considers only the gradient magnitude, while completely ignoring gradient orientation information, which plays a critical role in preserving fine structural details.

Some works have attempted to address this limitation by incorporating orientation-sensitive features into the fusion objective. For example, a loss function based on the Histogram of Oriented Gradients (HOG) has been introduced~\cite{long2023soft}, in which HOG, which encodes both gradient magnitude and orientation, is used as a supervisory signal and combined with a multi-scale structural similarity term to guide network training. To further mitigate ambiguous gradients, TC-MoA~\cite{zhu2024task} retains the sign of gradient values in all gradient-related terms. However, when producing gradients for each source image, it performs a per-pixel summation of the horizontal (x) and vertical (y) gradient responses, which causes mutual dilution and cancellation between directions, thereby weakening fine-detail preservation, as shown in Fig.~\ref{fig:gradloss_compare} (b). MRFS~\cite{zhang2024mrfs} decouples the Sobel gradients into x- and y- components, computes the maximum gradient in each direction across the two sources, and then enforces the fused image to match these maximum values along both axes. A major drawback of this approach is that when the two source images have inconsistent gradient polarities, the fusion result tends to favor the positive gradient even when the negative gradient has a larger magnitude.

To address the aforementioned issues, we propose a multi-scale direction-aligned gradient loss that fundamentally differs from prior scalar formulations (which rely only on magnitude or $\nabla_x+\nabla_y$). Instead, as shown in Fig.~\ref{fig:gradloss_compare} (c), our method directly matches the gradient vector ($\nabla_x, \nabla_y$) of the fused image to an axiswise, sign-preserving target selected from the visible and infrared sources.

This design offers several key advantages:
(i) it enforces orientation fidelity, since any incorrect edge direction is penalized along at least one axis;
(ii) it removes the directional bias inherent to collapsed directional derivatives (e.g., those computed along $(1,1)$ ), by treating the two axes symmetrically;
(iii) it enables fine-grained modality gating on a per-axis, per-pixel basis, allowing complementary contributions from each modality, for example, VIS for horizontal details and IR for vertical structures, particularly at corners and T-junctions; and
(iv) it improves training stability by aggregating supervision across multiple scales.

Designed to be simple yet effective, the proposed loss can serve as a plug-and-play replacement for traditional gradient losses in infrared-visible image fusion. When applied, it consistently produces fused images with sharper edges and more structurally consistent details across different spatial scales.

\section{Related work}\label{sec:related work}

\subsection{Infrared and visible image fusion}

The methods of using neural networks to fuse infrared and visible images can be roughly divided into the following four categories: generative models, models based on autoencoders, algorithm unrolling models and unified models~\cite{zhao2024emma}. Generative models represent the distribution of fused images and source images in the latent space. This includes two major categories, one is the earlier generative adversarial network method~\cite{xie2023r2f, ma2019fusiongan, ma2020ganmcc}, and the other is using denoising diffusion model that has emerged in the past two years~\cite{zhao2023ddfm, yiTextIFLeveragingSemantic2024}. The autoencoder-based methods use CNN/Transformer modules or a combination of the two as basic modules, obtain a powerful encoder in the training stage, adds some fusion rules in the prediction stage, and uses the decoder to generate a fused image~\cite{liang2022fusion,vs2022image}. For example, SDNet~\cite{zhang2021sdnet} incorporates an additional decomposition module that enhances detail retention in the fusion outcome by decomposing the fused image back into approximate source image pairs. CDDFuse~\cite{zhao2023cddfuse} proposed a feature decomposition fusion network that combines Transformer and CNN architectures with a correlation-driven loss function to effectively fuse modality-specific and shared features. 

The algorithm unrolling model shifts the focus of the algorithm from data-driven learning to model-driven learning, replaces complex manually designed operators with CNN/Transformer modules, and retains the original computational graph structure to achieve lightweight and interpretable learning~\cite{ju2022ivf,he2023degradation,zhao2021efficient}. A typical example is LRRNet~\cite{li2023lrrnet}, which uses optimization algorithms to guide the construction of model structures and uses the constructed model to solve algorithm optimization problems. It converts matrix multiplication into convolution operations, and the iterative optimization process is also replaced by a special feed forward network, thereby realizing an interpretable infrared and visible image fusion model. 

Unified models can break through the performance bottleneck of image fusion task by learning meta-knowledge across different tasks, and can also be deployed more quickly in practice. In addition, multimodal image fusion tasks are often integrated into coupled systems, including upstream (pre-processing) image registration~\cite{huang2022reconet, xu2023murf} and downstream object detection and semantic segmentation tasks~\cite{jiang2022towards}. Image registration can effectively remove image artifacts and misaligned areas, enhance edge clarity and expand perceptual range~\cite{wang2022unsupervised}. In addition, the gradient of recognition/detection loss in downstream tasks can effectively guide the generation of fused images. RFNet~\cite{xu2022rfnet} introduced an approach by combining image fusion task with image registration task, reinforcing each other rather than viewing them as separate tasks. Specifically, in the coarse registration phase, image-level metrics are established using image translation. In the fine registration phase, evaluation metrics are designed based on image fusion results. By adopting a coarse-to-fine approach for image registration, improved image registration results enhance image fusion performance. In turn, the outcomes of image fusion act as feedback to further refine the accuracy of image registration. SeAFusion~\cite{tang2022image} proposed a realtime semantic-aware image fusion framework. A semantic loss was introduced, which allows high-level semantic information to flow back to  image fusion module, to improve the facilitation of fused results for high-level vision tasks. Furthermore, some studies, such as GIFNet\cite{cheng2025one}, have proposed utilizing low-level fusion tasks, including multi-exposure and multi-focus fusion, as auxiliary supervision to guide the training of infrared–visible image fusion networks.

\subsection{Loss for IVIF}

There are generally three kinds of loss used in modern image fusion neural networks.

Basic per-pixel reconstruction losses like mean squared error (MSE)~\cite{fu2021dual} or mean absolute error (MAE) are often used to preserve overall brightness or intensity information. The fused image is encouraged to approximate the source images’ pixel values (or an appropriate composite, e.g. mean of source images~\cite{zhu2024task}) by minimizing pixel-wise error. A more modern approach is to calculate the per-pixel maximum of the two source images' grayscale values and then compute the L1 loss with the fused image~\cite{liu2025dcevo}. This helps ensure that the fused image does not deviate grossly in overall intensity or contrast from the inputs. In practice, some frameworks treat this as a reconstruction loss for each input image or for a weighted average of the inputs.

SSIM-based loss functions are widely adopted to maintain structural details from the source images. SSIM compares luminance, contrast, and structural similarity in local patches, correlating better with human perception than plain MSE. Many fusion methods replace or augment MSE with an SSIM loss to ensure the fused image has high structural similarity to the inputs~\cite{huang2022reconet}. In some cases a multi-scale SSIM variant is used to capture structures at multiple resolutions, further enhancing the preservation of both global structures and fine textures~\cite{long2023soft}.

To preserve edge details and texture from infrared and visible images, gradient-based losses are very common. These losses operate on high-frequency information: for instance, computing gradient maps (using Sobel filters, Laplacian operators, or Canny edges) for both the fused image and source images, and penalizing differences. This encourages the fused result to contain the strong edges from each modality. One approach applies a Laplacian filter to obtain gradient maps and then minimizes the MSE between the fused image’s gradients and the input image gradients~\cite{fu2021dual}. Another method extracts edge maps via the Canny operator and forces the fused image to mimic an edge-enhanced composite of the sources~\cite{lu2023fusion2fusion}. 

Some methods can directly generate a three-channel fused image, rather than first generating a single-channel fused image like most other methods, and then combining it with the chrominance channel of the visible light image to generate the final fused image. In order to avoid color distortion in the generated fused image, a type of color loss is introduced, such as calculating the l1 loss on the Cb and Cr chrominance channels~\cite{zhang2024mrfs}.

In practice, fusion frameworks combine the above loss terms in a weighted sum to form the final training objective.

\section{Method}\label{sec:method}


\subsection{Notation}

Let $I\in\mathbb{R}^{C \times H \times W}$ be an image. Let $\text{Sobel}_\text{x}$ and $\text{Sobel}_\text{y}$ be 2-D convolutions with Sobel kernels $K_x, K_y$. Define:
\begin{equation}
	K_x=\left[\begin{array}{ccc}
		-1 & 0 & 1 \\
		-2 & 0 & 2 \\
		-1 & 0 & 1
	\end{array}\right], \quad K_y=\left[\begin{array}{ccc}
		-1 & -2 & -1 \\
		0 & 0 & 0 \\
		1 & 2 & 1
	\end{array}\right]
\end{equation}Given an image $I$, define its horizontal and vertical gradients as

\begin{equation}
	\nabla_x I = \text{Sobel}_\text{x}(I), \quad
	\nabla_y I = \text{Sobel}_\text{y}(I)
\end{equation}and define the gradient magnitude as the sum of x- and y-components’ absolute values:

\begin{equation}
	\lVert\nabla I\rVert_1 \;\coloneqq\; \lVert\nabla_x I \rVert_1 + \lVert\nabla_y I\rVert_1
\end{equation}Let $I_\text{RGB}^{\text{vis}}\in\mathbb{R}^{C \times H \times W}$ be the visible image, $I^{\text{ir}}\in\mathbb{R}^{H \times W}$ the infrared image, and $I^f\in\mathbb{R}^{H \times W}$ the fused prediction. Usually, we convert the RGB visible image to YCrCb color space and only take the Y channel to do the subsequent processes, leave the other two channels just for imaging:

\begin{equation}
	\bigl(I^{\mathrm{vis}}_{Y},\, I^{\mathrm{vis}}_{Cr},\, I^{\mathrm{vis}}_{Cb}\bigr)
	= \operatorname{YCrCb} (I_\text{RGB}^{\mathrm{vis}})
\end{equation}

\subsection{Baseline gradient loss}

General gradient loss nowadays will be defined as follows:

\begin{equation}
	G^{\mathrm{vis}}=\left\|\nabla I_Y^{\mathrm{vis}}\right\|_1, \quad G^{\mathrm{ir}}=\left\|\nabla I^{\mathrm{ir}}\right\|_1, \quad G^f=\left\|\nabla I^f\right\|_1 
\end{equation}First calculate absolute gradient magnitude of three images respectively, then calculate element-wise hard maximum target:

\begin{equation}
	G^{\max} \;=\; \max\big(G^{\text{vis}}, G^{\text{ir}}\big)
\end{equation}finally calculate the mean absolute error:

\begin{equation}
	\begin{aligned}
		\mathcal{L}_{\text {grad }} &=\operatorname{MAE}\left(G^f, G^{\max }\right)\\&=
		\frac{1}{N H W} \sum_{n, h, w}\left|G_{n, h, w}^f-G_{n, h, w}^{\max }\right|
	\end{aligned}
\end{equation}This method collapses gradient to a scalar magnitude with direction or sign discarded.

\subsection{Signed gradient loss from TC-MoA} 

To avoid confusing gradients, TC-MoA retain the sign of the gradient values in all loss functions related to gradient information. It first calculate the signed gradient response without using $l_1$ norm

\begin{equation}\label{eq:tcmoa}
	G(I) \coloneqq \nabla_x I+\nabla_y I
\end{equation}For visible ( $I_Y^{\text {vis }}$ ) and infrared ( $I^{\text {ir }}$ ) sources,

\begin{equation}
	\begin{aligned}
		G^{\text{vis}} & =G\left(I_Y^{\mathrm{vis}}\right) \\
		G^{\text{ir}} & =G\left(I^{\mathrm{ir}}\right)
	\end{aligned}
\end{equation}Define an element-wise winner-take-all mask by comparing magnitudes of the signed responses:
\begin{equation}
	M=\mathbf{1}\left(\left|G^{\mathrm{vis}}\right| \geq\left|G^{\mathrm{ir}}\right|\right)
\end{equation}
so that selection is based on signal strength $|G(\cdot)|$ while the sign of the winner is retained. Form the sign-preserving target

\begin{equation}
	G^{\star}=M \odot G^{\mathrm{vis}}+(1-M) \odot G^{\mathrm{ir}} 
\end{equation}which keeps the full signed directional response from the dominant modality at each element (and each channel), thereby enforcing edge. For the fused prediction $I^f$, compute the fused directional response
\begin{equation}
	G^f=G\left(I^f\right)=\nabla_x I^f+\nabla_y I^f
\end{equation}The TC-MoA signed gradient loss is then computed as
\begin{equation}
	\mathcal{L}_{\text{grad}}=\operatorname{MAE}\left(G^f, G^{\star}\right)
\end{equation}However, equation \eqref{eq:tcmoa} projects the 2-D gradient vector onto the fixed direction $\mathbf{d}=(1,1)$. This collapses horizontal and vertical components into a single scalar and allows destructive interference between them. For example, if $\nabla_x I=a$ and $\nabla_y I=-a$ (a strong edge oriented near $\mathbf{d}_{\perp}=(1,-1)$ ), then $G(I)=0$ despite large edge energy $\|\nabla I\|_1=|a|+|a|=2|a|$. Hence, directions roughly orthogonal to $\mathbf{d}$ can be under-penalized, while structures aligned with $\mathbf{d}$ are over-emphasized, introducing an orientation anisotropy and potentially unstable gating when $|G(\cdot)|$ is used for winner selection. In practice, this projection may confuse gradient evidence from different directions (and even from different modalities at a pixel), attenuating true edges through cancellation and yielding an ambiguous supervision signal.

To avoid such cross-axis cancellation and directional bias, one should keep the gradient as a vector and supervise components separately, which is our design.

\subsection{Multi-scale gradient loss with direction alignment}

We seek to supervise the gradient vector of the fused image so that, at every location, each axis component ($x$ and $y$) inherits the dominant (in magnitude) and polarity-consistent signal from either visible or infrared images. Let $\mathcal{S}=\left\{s_1, \ldots, s_K\right\}$ be a set of image scales and $w_s>0$ their aggregation weights (normalized to $\sum_s w_s=1$ ).

For each $s \in \mathcal{S}$, we form resized images
\begin{equation}
	I_s^f=\mathcal{R}_s\left(I^f\right), \quad I_{Y, s}^{\mathrm{vis}}=\mathcal{R}_s\left(I_Y^{\mathrm{vis}}\right), \quad I_s^{\mathrm{ir}}=\mathcal{R}_s\left(I^{\mathrm{ir}}\right)
\end{equation}where $\mathcal{R}_s(\cdot)$ is bilinear resampling. This realizes a band-limited approximation of scale-space; gradients will therefore be compared at matched spatial bandwidths. Then we use Sobel operators with zero padding, we compute
\begin{equation}\label{eq:sobel}
	\begin{aligned}
		\left(\nabla_x^f, \nabla_y^f\right) & =\left(\mathcal{S}_x\left(I_s^f\right), \mathcal{S}_y\left(I_s^f\right)\right) \\
		\left(\nabla_x^{\mathrm{vis}}, \nabla_y^{\mathrm{vis}}\right) & =\left(\mathcal{S}_x\left(I_{Y, s}^{\mathrm{vis}}\right), \mathcal{S}_y\left(I_{Y, s}^{\mathrm{vis}}\right)\right)\\
		\left(\nabla_x^{\mathrm{ir}}, \nabla_y^{\mathrm{ir}}\right) & =\left(\mathcal{S}_x\left(I_s^{\mathrm{ir}}\right), \mathcal{S}_y\left(I_s^{\mathrm{ir}}\right)\right) 
	\end{aligned}
\end{equation}By retaining both components, we preserve edge orientation. For each axis we choose the modality with the larger absolute response, keeping its sign:
\begin{equation}
	M_x=\mathbf{1}\left(\left|\nabla_x^{\mathrm{vis}}\right| \geq\left|\nabla_x^{\mathrm{ir}}\right|\right), \quad M_y=\mathbf{1}\left(\left|\nabla_y^{\mathrm{vis}}\right| \geq\left|\nabla_y^{\mathrm{ir}}\right|\right)
\end{equation}
\begin{equation}
	\begin{aligned}
		\nabla_x^{\mathrm{sel}}&=M_x \nabla_x^{\mathrm{vis}}+\left(1-M_x\right) \nabla_x^{\mathrm{ir}}\\
		\nabla_y^{\mathrm{sel}}&=M_y \nabla_y^{\mathrm{vis}}+\left(1-M_y\right) \nabla_y^{\mathrm{ir}}
	\end{aligned}
\end{equation}This axis-wise gating allows the visible modality to dominate the horizontal gradient while the infrared modality dominates the vertical gradient at the same pixel and vice versa, which is crucial near corners and T-junctions. We penalize the $l_1$ distance between gradient vectors:
\begin{equation}
	\begin{aligned}
		\mathcal{L}_{s}&=\operatorname{MAE}\left(\nabla_x^f, \nabla_x^{\text {sel }}\right)+\operatorname{MAE}\left(\nabla_y^f, \nabla_y^{\text {sel }}\right) \\
		&=\left\|\left[\nabla_x^f, \nabla_y^f\right]-\left[\nabla_x^{\text {sel }}, \nabla_y^{\text {sel }}\right]\right\|_1
	\end{aligned}
\end{equation}Using $l_1$ (instead of $l_2$ ) reduces over-penalization of rare large discrepancies and empirically stabilizes training under outliers and aliasing at downsampled scales. The final objective sums across scales:
\begin{equation}
	\mathcal{L}_{\mathrm{grad}}=\sum_{s \in \mathcal{S}} w_s \mathcal{L}_s
\end{equation}We adopt equal weighting across all scales (for example, $w_s=1 /len(\mathcal{S})$,  $\mathcal{S}=\{1,0.5,0.25\}$ ) to ensure that both fine and coarse structures contribute uniformly to the supervision. In \eqref{eq:sobel}, zero padding is applied for convolution, as it provides a neutral boundary condition that avoids introducing artificial edges and yields cleaner fusion boundaries.

%

\section{Experiment results}\label{sec:Experiment results}

\subsection{Evaluation metrics}Below we briefly describe the evaluation criteria used. Single-image metrics: EN~\cite{roberts2008en} (entropy, $\uparrow$) quantifies information content;  SD~\cite{eskicioglu1995sd} (standard deviation, $\uparrow$) captures global contrast.

Reference-based metrics (computed between the fused image and the source images, then averaged when applicable): MI (mutual information, $\uparrow$) gauges information preserved from the sources;  SCD~\cite{aslantas2015scd} (sum of correlations of differences, $\uparrow$) assesses structural/detail transfer via correlations of high-frequency differences; VIF~\cite{han2013vif} (visual information fidelity, $\uparrow$) estimates perceptual fidelity under an HVS model; $Q_{A B / F}$ (edge-based fusion quality, $\uparrow$) evaluates edge/structure preservation. See~\cite{yang2024review} for the calculation formulas and detailed explanations of all metrics.

\subsection{MSRS results}
\label{sec:MSRS results}

We adopt the open-source ReCoNet model~\cite{huang2022reconet} as our baseline. To isolate the effect of the fusion objective, we disable the built-in image calibration module and train/use only the fusion module. We compare different loss formulations by training for 300 epochs on the MSRS\footnote{\url{https://github.com/Linfeng-Tang/MSRS}} training set (1,083 image pairs, resolution $640 \times 480$), using a random 4:1 split for training and validation. Evaluation is conducted on the original MSRS test set comprising 361 pairs. The Adam optimizer updates the parameters with the learning rate of 0.001. As summarized in table~\ref{tab:msrs}, the \textit{ori} configuration employs the model’s original SSIM-based structural loss and intensity reconstruction loss with a weighting of 3:7 (closely following the original paper). The \textit{grad} configuration adds a conventional gradient loss to the above, with weights 1.5:7:1.5 for (SSIM, intensity, gradient); because the gradient term is also a structural constraint, we preserve the overall structural-to-intensity weighting at 3:7. The \textit{tcmoa} configuration replaces the conventional gradient term with the directional gradient loss used in TC-MoA. Finally, \textit{ours} substitutes our proposed gradient loss for the conventional gradient term, using the same 1.5:7:1.5 weighting.

\begin{figure*}[ht]
	\centering
	\includegraphics[width=0.98\textwidth]{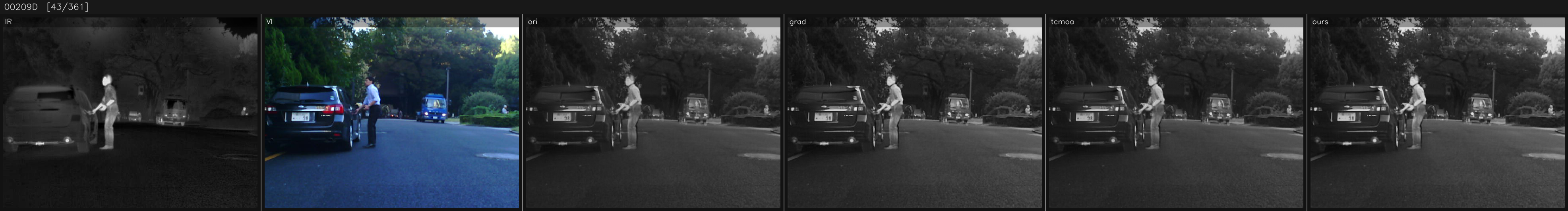}
	\vspace{-4pt}
	\includegraphics[width=0.98\textwidth]{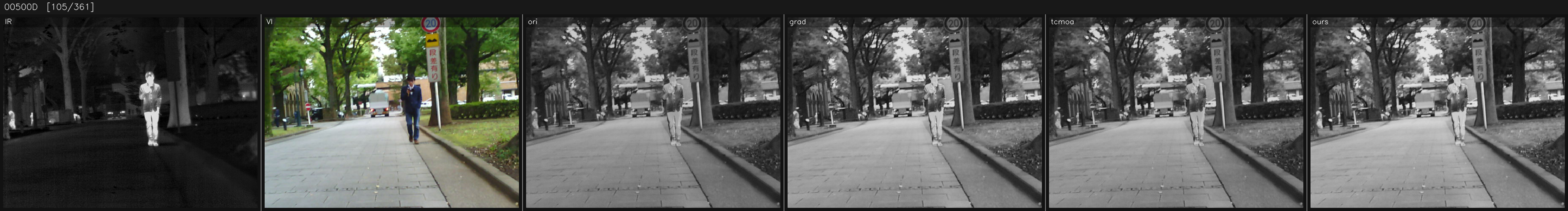}
	\caption{Qualitative comparison on the MSRS dataset. From left to right: (a) infrared image, (b) visible image, (c) ORI, (d) GRAD, (e) TCMoA, and (f) Ours.}
	\label{fig:msrs-qual}
\end{figure*}

Figure~\ref{fig:msrs-qual} presents representative qualitative results on the MSRS dataset.
Visually, the fused images produced by the \textit{Grad} and \textit{Ours} models exhibit pedestrians that appear more distinct and visually salient compared to those generated by the \textit{ORI} and \textit{TCMoA} groups.
This observation suggests that gradient-based objectives contribute positively to preserving local contrast and enhancing the visibility of salient targets.
In particular, the darker appearance of the \textit{TCMoA} results can be attributed to gradient cancellation: when the gradients along the $x$- and $y$-directions have opposite signs, their combination suppresses the overall edge magnitude, thereby reducing perceived brightness in high-contrast regions.
Traditional gradient losses typically sum the gradient magnitudes from both directions, while our method independently selects the maximum magnitude along each direction.
This strategy effectively prevents mutual cancellation and preserves stronger directional gradients, leading to higher local contrast and visually brighter fusion outputs.
Overall, these findings are fully consistent with our theoretical expectations that a more balanced gradient constraint yields clearer edges and more salient target regions.

\begin{table}[ht]
	\centering
	\caption{Quantitative results of ReCoNet trained with different loss functions on the MSRS dataset. All metrics are averaged over the test set and rounded to three decimal places. The best results are in \textcolor{red}{red} and the second best in \textcolor{blue}{blue}.}
	\label{tab:msrs}
	\begin{tabular}{lrrrrrr}
		\toprule
		& EN $\uparrow$ & MI $\uparrow$ & SD $\uparrow$ & SCD $\uparrow$ & VIF $\uparrow$ & $Q_{A B / F}$ $\uparrow$\\
		\midrule
		ori        & 6.188 & 3.092 & 33.284 & 1.436 & 0.761 & 0.539\\
		grad   & 6.402 & 3.429 & 38.428 & 1.541 & \textcolor{blue}{0.842} & \textcolor{red}{0.607}\\
		tcmoa  & \textcolor{blue}{6.413} & \textcolor{blue}{3.471} & \textcolor{blue}{39.043} & \textcolor{blue}{1.596} & \textcolor{blue}{0.842} & \textcolor{blue}{0.602}\\
		ours & \textcolor{red}{6.447} & \textcolor{red}{3.552} & \textcolor{red}{39.744} & \textcolor{red}{1.603} & \textcolor{red}{0.851} & \textcolor{red}{0.607}\\
		\bottomrule
	\end{tabular}
\end{table}

As summarized in Table~\ref{tab:msrs}, the proposed direction-aware gradient fusion loss yields consistent improvements across both information-oriented and perceptual/structural metrics.
On the MSRS dataset, \textit{ours} achieves the best scores on all six evaluated metrics---EN, MI, SD, SCD, VIF, and $Q_{A B / F}$---while \textit{tcmoa} generally ranks second.
The baseline configuration (\textit{ori}), which relies solely on SSIM and intensity reconstruction, lacks explicit structural regularization and thus produces comparatively lower SD, SCD, and VIF values.
In contrast, the conventional gradient loss (\textit{grad}) enhances texture representation but remains limited in handling directionally inconsistent gradients. \textit{tcmoa} alleviates this issue partially through directional filtering, yet its hard-max operation on $|G|$ causes pixels with strong but cross-oriented gradients to be under-supervised, resulting in slightly weaker SCD, VIF, and $Q_{A B / F}$ than our formulation.

Notably, replacing part of the SSIM loss with gradient-based constraints also leads to clear gains in EN and MI. This suggests that the gradient losses not only strengthen structural fidelity but also promote richer information content in the fused images. By enhancing sensitivity to edge and texture variations, the gradient constraints effectively mitigate over-smoothing caused by excessive structural regularization from SSIM. Consequently, the fused outputs preserve more spatial detail and complementary modality information, reflected by higher entropy (EN) and mutual information (MI) values.
Among all variants, \textit{ours} achieves the highest EN and MI, demonstrating that it better balances contrast enhancement and information retention without introducing visible artifacts.

Quantitatively, \textit{ours} surpasses \textit{tcmoa} by approximately $0.4{-}2.3\%$ and \textit{grad} by $0.4{-}3.4\%$ across EN, MI, SD, SCD, VIF, and $Q_{A B / F}$. Overall, the proposed loss effectively reinforces structural fidelity, perceptual consistency, and information richness while maintaining stable optimization dynamics throughout training.

\subsection{FMB results}
\label{sec:FMB results}

To further assess the robustness of our gradient alignment loss, we evaluated the models trained on MSRS dataset directly on 280 image pairs from the FMB test set, without any additional fine-tuning. As shown in Table~\ref{tab:msrs2fmb}, \textit{ours} achieves the best results across all metrics, while \textit{grad} generally ranks second. Interestingly, \textit{tcmoa} performs worse than the conventional gradient loss, suggesting that the FMB dataset contains more cases where horizontal and vertical gradients are not well aligned.

Quantitatively, \textit{ours} consistently improves both low-level statistics (EN, SD) and perceptual metrics (VIF, $Q_{AB/F}$), demonstrating that the proposed loss enhances not only contrast and texture richness but also cross-modality structural fidelity. The concurrent increase in MI and SCD further indicates that the fused representations contain more complementary, non-redundant information from both modalities.

The inferior performance of \textit{cmoa} can be attributed to its sign-sensitive and direction-coupled gradient formulation. Specifically, the SobelxyRGB operator in tcmoa computes the signed sum of horizontal and vertical gradients, for example, $G = G_x + G_y$. When $G_x$ and $G_y$ have opposite signs---as frequently occurs at diagonal or texture-rich boundaries---their magnitudes partially cancel out, leading to an underestimated overall gradient strength. This reduction directly weakens the subsequent magnitude-based masking in MaxGradLoss, causing the model to lose edge information in regions where the two axes are not aligned.

In contrast, our proposed loss decouples the two directions and performs axis-wise dominant selection rather than signed aggregation. By taking the stronger response between modalities separately for $G_x$ and $G_y$, and by averaging across scales, our formulation avoids destructive interference between orthogonal gradients. This design preserves true gradient intensity even when the edge orientation varies, which explains the improved robustness of \textit{ours} on the FMB dataset where such misaligned gradients are more prevalent.

Overall, these results highlight that enforcing direction-consistent, scale-adaptive gradient alignment provides a stronger inductive bias for multimodal fusion. The proposed loss not only maintains detail fidelity within the training domain but also generalizes effectively to unseen datasets with distinct gradient characteristics, thereby improving both quantitative metrics and perceptual visual quality.

\begin{table}[ht]
	\centering
	\caption{Quantitative results on the FMB dataset. All metrics are averaged over the test set and rounded to three decimal places. 
		The best results are in \textcolor{red}{red} and the second best in \textcolor{blue}{blue}.}
	\label{tab:msrs2fmb}
	\begin{tabular}{lrrrrrr}
		\toprule
		& EN $\uparrow$ & MI $\uparrow$ & SD $\uparrow$ & SCD $\uparrow$ & VIF $\uparrow$ & $Q_{A B / F}$ $\uparrow$\\
		\midrule
		ori        & 6.437 & 3.379 & 27.820 & 1.459 & 0.686 & 0.556\\
		grad   & \textcolor{blue}{6.549} & \textcolor{blue}{3.589} & \textcolor{blue}{31.819} & 1.460 & \textcolor{blue}{0.743} & \textcolor{blue}{0.620}\\
		tcmoa  & 6.486 & 3.479 & 29.520 & \textcolor{blue}{1.479} & 0.715 & 0.593\\
		ours & \textcolor{red}{6.578} & \textcolor{red}{3.684} & \textcolor{red}{32.567} & \textcolor{red}{1.511} & \textcolor{red}{0.762} & \textcolor{red}{0.626}\\
		\bottomrule
	\end{tabular}
\end{table}

\subsection{M3FD results}
\label{sec:M3FD results}

To further verify cross-dataset robustness, we evaluated the same models on 300 image pairs from the M3FD fusion dataset~\cite{liu2022target}, using the networks trained on MSRS without additional fine-tuning. As shown in Table~\ref{tab:msrs2m3fd}, the performance ranking remains consistent with the previous FMB results, where \textit{ours} achieves the best overall results and \textit{grad} ranks second. This consistent trend demonstrates that the proposed gradient alignment loss generalizes well to datasets with different acquisition settings and illumination conditions.

Notably, the SCD score of our method is slightly lower than that of the original configuration. This can be attributed to the inherent nature of SCD, which tends to reward independent variations between the fused and source images, even when they arise from redundant or noisy details. By enforcing axis-wise and multi-scale gradient alignment, our formulation suppresses such artifacts and yields cleaner, more coherent edges. As a result, while SCD marginally decreases, other information-based and perceptual metrics (EN, MI, VIF, $Q_{AB/F}$) consistently improve, indicating a more balanced and visually faithful fusion outcome.

\begin{table}[ht]
	\centering
	\caption{Quantitative results on the M3FD dataset. All metrics are averaged over the test set and rounded to three decimal places. 
		The best results are in \textcolor{red}{red} and the second best in \textcolor{blue}{blue}.}
	\label{tab:msrs2m3fd}
	\begin{tabular}{lrrrrrr}
		\toprule
		& EN $\uparrow$ & MI $\uparrow$ & SD $\uparrow$ & SCD $\uparrow$ & VIF $\uparrow$ & $Q_{A B / F}$ $\uparrow$\\
		\midrule
		ori        & 6.498 & 3.146 & 26.806 & \textcolor{red}{1.502} & 0.629 & 0.485\\
		grad   & \textcolor{blue}{6.681} & \textcolor{blue}{3.454} & \textcolor{blue}{31.291} & 1.439 & \textcolor{blue}{0.674} & \textcolor{blue}{0.555}\\
		tcmoa  & 6.582 & 3.351 & 28.704 & 1.494 & 0.653 & 0.525\\
		ours & \textcolor{red}{6.707} & \textcolor{red}{3.562} & \textcolor{red}{32.258} & \textcolor{blue}{1.496} & \textcolor{red}{0.692} & \textcolor{red}{0.559}\\
		\bottomrule
	\end{tabular}
\end{table}

\subsection{LLVIP results}

To further examine scalability and stability under large-scale training, we conducted experiments on the LLVIP dataset~\cite{jiaLLVIPVisibleinfraredPaired2023}, which contains over 12k training and 3k testing pairs, significantly larger than other benchmarks. All models were retrained for 300 epochs using the same settings as in Section~\ref{sec:MSRS results}. As reported in Table~\ref{tab:llvip}, \textit{ours} achieves the best performance in most metrics, especially in EN, MI, SD, and SCD, demonstrating that the proposed gradient alignment loss continues to enhance detail richness and structural preservation even when trained on large-scale data. These gains reflect that the proposed direction-aligned gradient loss effectively enhances information transfer and structural contrast under challenging low-light conditions. Specifically, the higher MI and EN values indicate that the fused images retain more complementary information from both modalities, while the increased SD and SCD confirm stronger global contrast and more complete structural integration. Such results demonstrate that our multi-scale, axis-wise alignment promotes sharper, more coherent edge representations that are better suited for heterogeneous illumination and texture characteristics in LLVIP.

However, our method obtains a slightly lower $Q_{AB/F}$ value compared with the conventional gradient loss. This difference arises from the inherent property of $Q_{AB/F}$, which measures the local gradient correlation between the fused and source images. The metric favors strict directional consistency with the original gradients, rewarding edge preservation even when those edges are weak, noisy, or misaligned across modalities. In contrast, our loss explicitly encourages consistent edge orientation and magnitude between modalities rather than mere reproduction of the original gradient field. During training, this mechanism suppresses redundant or ambiguous gradients and reconstructs missing boundaries in poorly illuminated regions—actions that improve perceptual sharpness and information fidelity but inevitably reduce gradient correlation as measured by $Q_{AB/F}$.

Our method shows a slight decrease in VIF compared with the conventional gradient-based loss. This observation can be attributed to the nature of the VIF metric, which measures the statistical similarity of visual information between the fused and source images. Since VIF is derived from natural scene statistics, it favors results that closely preserve the original intensity and structural distributions. Our direction-aligned gradient loss, however, enhances edge magnitude and contrast by realigning gradient orientations across modalities. Although this process slightly alters the original image statistics, leading to a minor reduction in VIF, it contributes to sharper, more perceptually distinct fusion results with richer detail and improved information content.

Overall, the observed trade-off highlights a fundamental distinction between perceptual quality and gradient similarity. While VIF and $Q_{AB/F}$ slightly decrease, the considerable improvements in information-based (MI, EN) and structural (SD, SCD) metrics indicate that our method achieves a more balanced and visually meaningful fusion, preserving salient structures while avoiding the amplification of redundant or noisy edges.

\begin{table}[ht]
	\centering
	\caption{Quantitative results on the LLVIP dataset. All metrics are averaged over the test set and rounded to three decimal places. 
		The best results are in \textcolor{red}{red} and the second best in \textcolor{blue}{blue}.}
	\label{tab:llvip}
	\begin{tabular}{lrrrrrr}
		\toprule
		& EN $\uparrow$ & MI $\uparrow$ & SD $\uparrow$ & SCD $\uparrow$ & VIF $\uparrow$ & $Q_{A B / F}$ $\uparrow$\\
		\midrule
		ori        & 7.104 & 3.467 & 41.567 & 1.384 & 0.745 & 0.495\\
		grad   & \textcolor{blue}{7.181} & 3.536 & \textcolor{blue}{43.588} & 1.345 & \textcolor{red}{0.832} & \textcolor{red}{0.646}\\
		tcmoa  & 7.144 & \textcolor{blue}{3.637} & 42.849 & \textcolor{blue}{1.393} & 0.783 & 0.573\\
		ours & \textcolor{red}{7.263} & \textcolor{red}{3.909} & \textcolor{red}{46.252} & \textcolor{red}{1.461} & \textcolor{blue}{0.826} & \textcolor{blue}{0.590}\\
		\bottomrule
	\end{tabular}
\end{table}

\subsection{Ablation study}
\label{sec:Ablation study}

The ablation study on the gradient fusion method is not repeated here, as it largely overlaps with the previous comparative experiments. Instead, we investigate the effects of other design variations in the proposed loss function, including:
\begin{enumerate}
	\item Loss I: single-scale version;
	\item Loss II: multi-scale version with non-uniform weights (1, 0.618, 0.618\textsuperscript{2}, normalized to 0.500, 0.309, 0.191);
	\item Loss III: same weighting across scales but with reflect padding in sobel operations;
	\item Loss IV (ours): multi-scale with equal weights and zero padding in sobel operations.
\end{enumerate}All other training settings are identical to those described in Section~\ref{sec:MSRS results}, except that the gradient loss is replaced with the above variants. The corresponding quantitative results are summarized in Table~\ref{tab:ablation}.

\begin{table}[ht]
	\centering
	\caption{Ablation study on MSRS dataset. All metrics are averaged over the test set and rounded to three decimal places. 
		The best results are in \textcolor{red}{red} and the second best in \textcolor{blue}{blue}.}
	\label{tab:ablation}
	\begin{tabular}{lrrrrrr}
		\toprule
		Loss & EN $\uparrow$ & MI $\uparrow$ & SD $\uparrow$ & SCD $\uparrow$ & VIF $\uparrow$ & $Q_{A B / F}$ $\uparrow$\\
		\midrule
		\text{I.}    & 6.366 & 3.351 & 37.832 & 1.566 & 0.830 & \textcolor{red}{0.602} \\
		\text{II.}   & \textcolor{blue}{6.428} & 3.504 & 39.368 & 1.597 & \textcolor{blue}{0.847} & \textcolor{red}{0.602} \\
		\text{III.}  & 6.426 & \textcolor{blue}{3.509} & \textcolor{blue}{39.389} & \textcolor{red}{1.604} & 0.846 & 0.600 \\
		\text{IV.} & \textcolor{red}{6.447} & \textcolor{red}{3.552} & \textcolor{red}{39.744} & \textcolor{blue}{1.603} & \textcolor{red}{0.851} & \textcolor{blue}{0.601} \\
		\bottomrule
	\end{tabular}
\end{table}

As shown in Table~\ref{tab:ablation}, among all configurations, incorporating multi-scale gradient alignment yields the most significant improvement, achieving approximately 1.2-6\% gains across the first five metrics compared with the single-scale version. This confirms that introducing multiple spatial scales helps the network capture both fine and coarse structural cues, while mitigating the influence of local noise. The coarse-scale gradients provide stable, low-frequency guidance that complements fine-scale edge details, leading to better overall structure preservation and contrast enhancement.

In Loss II, we applied non-uniform scale weights following the golden ratio (1:0.618:0.618\textsuperscript{2}). Intuitively, larger weights at finer scales should emphasize local texture details, while smaller weights at coarser scales help regularize global structure. However, the experimental results reveal that equal weighting (Loss IV) performs slightly better. This suggests that gradient information at different spatial scales contributes comparably to the final fusion quality, and overemphasizing finer scales may lead to redundant or noisy edge reinforcement. While alternative weighting schemes could potentially yield further gains, we leave this exploration for future work.

For Loss III, we replaced zero padding with reflect padding to reduce potential edge attenuation caused by zero-valued boundaries. Surprisingly, this modification did not lead to improved performance. In fact, zero padding (Loss IV) produced slightly higher scores on most metrics. We attribute this to the fact that reflect padding introduces artificial symmetry near the image borders, which can create inconsistent gradient patterns when fusing heterogeneous modalities. Zero padding, on the other hand, provides a more neutral boundary condition that avoids such interference.

Overall, the final version (Loss IV), featuring multi-scale alignment with equal weights and zero padding, achieves the most consistent and stable performance across all metrics. Although it trails slightly in SCD and $Q_{AB/F}$, its superior results in information-based and perceptual measures (EN, MI, SD, VIF) indicate a more balanced and robust fusion behavior.

\section{Conclusion}\label{sec:conclusion}

In this paper, we have introduced a plug-and-play gradient loss for infrared and visible image fusion. The proposed loss addresses two long-standing limitations of conventional gradient-based fusion objectives: (1) most existing formulations focus solely on gradient magnitude while ignoring directional consistency, and (2) direction-aware variants often suffer from gradient cancellation when opposite directions coexist. Our method achieves direction-consistent alignment by explicitly enforcing axis-wise and multi-scale gradient consistency, effectively preserving both edge magnitude and orientation.

Extensive experiments on four publicly available datasets demonstrate that the proposed loss can be seamlessly integrated into existing fusion networks and consistently improves performance over traditional gradient losses. The ablation studies further confirm the effectiveness of our design choices, including the use of multi-scale alignment, equal weighting, and zero-padding strategies.

Although the method proves robust and generalizable, we have not yet conducted an exhaustive exploration of certain hyperparameters, such as the number of scales, the choice of scale ratios, and the inter-scale weighting strategy. We believe that a more systematic investigation of these factors may yield further improvements. Future work will also explore extending the proposed direction-consistent loss to other multimodal fusion tasks and downstream perception-oriented objectives.

\section*{Declaration of interests}
The authors declare that they have no known competing financial interests or personal relationships that could have appeared to influence the work reported in this paper.

\section*{Data availability}
The code will be made public in the near future.

\section*{Acknowledgements}
This research was funded by [Infrared vision theory and method] grant number [2023JCJQ-ZD-011-12], and the APC was funded by [Infrared vision theory and method].



\bibliographystyle{elsarticle-num} 
\bibliography{bibliography}

\begin{thebibliography}{10}
\expandafter\ifx\csname url\endcsname\relax
  \def\url#1{\texttt{#1}}\fi
\expandafter\ifx\csname urlprefix\endcsname\relax\def\urlprefix{URL }\fi
\expandafter\ifx\csname href\endcsname\relax
  \def\href#1#2{#2} \def\path#1{#1}\fi

\bibitem{yang2024lfdt}
B.~Yang, Z.~Jiang, D.~Pan, H.~Yu, G.~Gui, W.~Gui, Lfdt-fusion: A latent
  feature-guided diffusion transformer model for general image fusion,
  Information Fusion (2024) 102639.

\bibitem{yang2024review}
K.~Yang, W.~Xiang, Z.~Chen, J.~Zhang, Y.~Liu, A review on infrared and visible
  image fusion algorithms based on neural networks, Journal of Visual
  Communication and Image Representation (2024) 104179\href
  {https://doi.org/https://doi.org/10.1016/j.jvcir.2024.104179}
  {\path{doi:https://doi.org/10.1016/j.jvcir.2024.104179}}.

\bibitem{yan2022learning}
Z.~Yan, K.~Wang, X.~Li, Z.~Zhang, G.~Li, J.~Li, J.~Yang, Learning complementary
  correlations for depth super-resolution with incomplete data in real world,
  IEEE transactions on neural networks and learning systems 35~(4) (2022)
  5616--5626.

\bibitem{yan2022rignet}
Z.~Yan, K.~Wang, X.~Li, Z.~Zhang, J.~Li, J.~Yang, Rignet: Repetitive image
  guided network for depth completion, in: European Conference on Computer
  Vision, Springer, 2022, pp. 214--230.

\bibitem{zhao2022discrete}
Z.~Zhao, J.~Zhang, S.~Xu, Z.~Lin, H.~Pfister, Discrete cosine transform network
  for guided depth map super-resolution, in: Proceedings of the IEEE/CVF
  conference on computer vision and pattern recognition, 2022, pp. 5697--5707.

\bibitem{jiang2022towards}
Z.~Jiang, Z.~Zhang, X.~Fan, R.~Liu, Towards all weather and unobstructed
  multi-spectral image stitching: Algorithm and benchmark, in: Proceedings of
  the 30th ACM international conference on multimedia, 2022, pp. 3783--3791.

\bibitem{wang2022unsupervised}
D.~Wang, J.~Liu, X.~Fan, R.~Liu, Unsupervised misaligned infrared and visible
  image fusion via cross-modality image generation and registration, arXiv
  preprint arXiv:2205.11876 (2022).

\bibitem{xu2022rfnet}
H.~Xu, J.~Ma, J.~Yuan, Z.~Le, W.~Liu, Rfnet: Unsupervised network for mutually
  reinforcing multi-modal image registration and fusion, in: Proceedings of the
  IEEE/CVF conference on computer vision and pattern recognition, 2022, pp.
  19679--19688.

\bibitem{liu2022target}
J.~Liu, X.~Fan, Z.~Huang, G.~Wu, R.~Liu, W.~Zhong, Z.~Luo, Target-aware dual
  adversarial learning and a multi-scenario multi-modality benchmark to fuse
  infrared and visible for object detection, in: Proceedings of the IEEE/CVF
  Conference on Computer Vision and Pattern Recognition, 2022, pp. 5802--5811.

\bibitem{tang2022image}
L.~Tang, J.~Yuan, J.~Ma, Image fusion in the loop of high-level vision tasks: A
  semantic-aware real-time infrared and visible image fusion network,
  Information Fusion 82 (2022) 28--42.

\bibitem{zhangRethinkingImageFusion2020}
H.~Zhang, H.~Xu, Y.~Xiao, X.~Guo, J.~Ma, Rethinking the {{Image Fusion}}: {{A
  Fast Unified Image Fusion Network}} based on {{Proportional Maintenance}} of
  {{Gradient}} and {{Intensity}}, Proceedings of the AAAI Conference on
  Artificial Intelligence 34~(07) (2020) 12797--12804.
\newblock \href {https://doi.org/10.1609/aaai.v34i07.6975}
  {\path{doi:10.1609/aaai.v34i07.6975}}.

\bibitem{zhao2023cddfuse}
Z.~Zhao, H.~Bai, J.~Zhang, Y.~Zhang, S.~Xu, Z.~Lin, R.~Timofte, L.~Van~Gool,
  Cddfuse: Correlation-driven dual-branch feature decomposition for
  multi-modality image fusion, in: Proceedings of the IEEE/CVF Conference on
  Computer Vision and Pattern Recognition, 2023, pp. 5906--5916.

\bibitem{long2023soft}
Y.~Long, W.~Lai, H.~Zhang, et~al., Soft histogram of gradients loss: A loss
  function for optimization of the image fusion networks, Laser \&
  Optoelectronics Progress 60~(24) (2023) 2411001.

\bibitem{zhu2024task}
P.~Zhu, Y.~Sun, B.~Cao, Q.~Hu, Task-customized mixture of adapters for general
  image fusion, in: Proceedings of the IEEE/CVF conference on computer vision
  and pattern recognition, 2024, pp. 7099--7108.

\bibitem{zhang2024mrfs}
H.~Zhang, X.~Zuo, J.~Jiang, C.~Guo, J.~Ma, Mrfs: Mutually reinforcing image
  fusion and segmentation, in: Proceedings of the IEEE/CVF Conference on
  Computer Vision and Pattern Recognition, 2024, pp. 26974--26983.

\bibitem{zhao2024emma}
Z.~Zhao, H.~Bai, J.~Zhang, Y.~Zhang, K.~Zhang, S.~Xu, D.~Chen, R.~Timofte,
  L.~Van~Gool, Equivariant multi-modality image fusion, in: Proceedings of the
  IEEE/CVF Conference on Computer Vision and Pattern Recognition, 2024, pp.
  25912--25921.

\bibitem{xie2023r2f}
Y.~Xie, G.~Liu, R.~Xu, D.~P. Bavirisetti, H.~Tang, M.~Xing, R2f-ugcgan: a
  regional fusion factor-based union gradient and contrast generative
  adversarial network for infrared and visible image fusion, Journal of Modern
  Optics 70~(1) (2023) 52--68.

\bibitem{ma2019fusiongan}
J.~Ma, W.~Yu, P.~Liang, C.~Li, J.~Jiang, Fusiongan: A generative adversarial
  network for infrared and visible image fusion, Information fusion 48 (2019)
  11--26.

\bibitem{ma2020ganmcc}
J.~Ma, H.~Zhang, Z.~Shao, P.~Liang, H.~Xu, Ganmcc: A generative adversarial
  network with multiclassification constraints for infrared and visible image
  fusion, IEEE Transactions on Instrumentation and Measurement 70 (2020) 1--14.

\bibitem{zhao2023ddfm}
Z.~Zhao, H.~Bai, Y.~Zhu, J.~Zhang, S.~Xu, Y.~Zhang, K.~Zhang, D.~Meng,
  R.~Timofte, L.~Van~Gool, Ddfm: denoising diffusion model for multi-modality
  image fusion, in: Proceedings of the IEEE/CVF International Conference on
  Computer Vision, 2023, pp. 8082--8093.

\bibitem{yiTextIFLeveragingSemantic2024}
X.~Yi, H.~Xu, H.~Zhang, L.~Tang, J.~Ma, Text-{{IF}}: {{Leveraging Semantic Text
  Guidance}} for {{Degradation-Aware}} and {{Interactive Image Fusion}} (Mar.
  2024).
\newblock \href {http://arxiv.org/abs/2403.16387} {\path{arXiv:2403.16387}},
  \href {https://doi.org/10.48550/arXiv.2403.16387}
  {\path{doi:10.48550/arXiv.2403.16387}}.

\bibitem{liang2022fusion}
P.~Liang, J.~Jiang, X.~Liu, J.~Ma, Fusion from decomposition: A self-supervised
  decomposition approach for image fusion, in: European Conference on Computer
  Vision, Springer, 2022, pp. 719--735.

\bibitem{vs2022image}
V.~Vs, J.~M.~J. Valanarasu, P.~Oza, V.~M. Patel, Image fusion transformer, in:
  2022 IEEE International conference on image processing (ICIP), IEEE, 2022,
  pp. 3566--3570.

\bibitem{zhang2021sdnet}
H.~Zhang, J.~Ma, Sdnet: A versatile squeeze-and-decomposition network for
  real-time image fusion, International Journal of Computer Vision 129 (2021)
  2761--2785.

\bibitem{ju2022ivf}
M.~Ju, C.~He, J.~Liu, B.~Kang, J.~Su, D.~Zhang, Ivf-net: An infrared and
  visible data fusion deep network for traffic object enhancement in
  intelligent transportation systems, IEEE Transactions on Intelligent
  Transportation Systems 24~(1) (2022) 1220--1234.

\bibitem{he2023degradation}
C.~He, K.~Li, G.~Xu, Y.~Zhang, R.~Hu, Z.~Guo, X.~Li, Degradation-resistant
  unfolding network for heterogeneous image fusion, in: Proceedings of the
  IEEE/CVF international conference on computer vision, 2023, pp. 12611--12621.

\bibitem{zhao2021efficient}
Z.~Zhao, S.~Xu, J.~Zhang, C.~Liang, C.~Zhang, J.~Liu, Efficient and model-based
  infrared and visible image fusion via algorithm unrolling, IEEE Transactions
  on Circuits and Systems for Video Technology 32~(3) (2021) 1186--1196.

\bibitem{li2023lrrnet}
H.~Li, T.~Xu, X.-J. Wu, J.~Lu, J.~Kittler, Lrrnet: A novel representation
  learning guided fusion network for infrared and visible images, IEEE
  transactions on pattern analysis and machine intelligence (2023).

\bibitem{huang2022reconet}
Z.~Huang, J.~Liu, X.~Fan, R.~Liu, W.~Zhong, Z.~Luo, Reconet: Recurrent
  correction network for fast and efficient multi-modality image fusion, in:
  European Conference on Computer Vision, Springer, 2022, pp. 539--555.

\bibitem{xu2023murf}
H.~Xu, J.~Yuan, J.~Ma, Murf: Mutually reinforcing multi-modal image
  registration and fusion, IEEE transactions on pattern analysis and machine
  intelligence 45~(10) (2023) 12148--12166.

\bibitem{cheng2025one}
C.~Cheng, T.~Xu, Z.~Feng, X.~Wu, Z.~Tang, H.~Li, Z.~Zhang, S.~Atito, M.~Awais,
  J.~Kittler, One model for all: Low-level task interaction is a key to
  task-agnostic image fusion, in: Proceedings of the Computer Vision and
  Pattern Recognition Conference, 2025, pp. 28102--28112.

\bibitem{fu2021dual}
Y.~Fu, X.-J. Wu, A dual-branch network for infrared and visible image fusion,
  in: 2020 25th international conference on pattern recognition (ICPR), IEEE,
  2021, pp. 10675--10680.

\bibitem{liu2025dcevo}
J.~Liu, B.~Zhang, Q.~Mei, X.~Li, Y.~Zou, Z.~Jiang, L.~Ma, R.~Liu, X.~Fan,
  Dcevo: Discriminative cross-dimensional evolutionary learning for infrared
  and visible image fusion, in: Proceedings of the Computer Vision and Pattern
  Recognition Conference, 2025, pp. 2226--2235.

\bibitem{lu2023fusion2fusion}
C.~Lu, H.~Qin, Z.~Deng, Z.~Zhu, Fusion2fusion: An infrared--visible image
  fusion algorithm for surface water environments, Journal of Marine Science
  and Engineering 11~(5) (2023) 902.

\bibitem{roberts2008en}
J.~W. Roberts, J.~A. Van~Aardt, F.~B. Ahmed, Assessment of image fusion
  procedures using entropy, image quality, and multispectral classification,
  Journal of Applied Remote Sensing 2~(1) (2008) 023522.

\bibitem{eskicioglu1995sd}
A.~M. Eskicioglu, P.~S. Fisher, Image quality measures and their performance,
  IEEE Transactions on communications 43~(12) (1995) 2959--2965.

\bibitem{aslantas2015scd}
V.~Aslantas, E.~Bendes, A new image quality metric for image fusion: The sum of
  the correlations of differences, Aeu-international Journal of electronics and
  communications 69~(12) (2015) 1890--1896.

\bibitem{han2013vif}
Y.~Han, Y.~Cai, Y.~Cao, X.~Xu, A new image fusion performance metric based on
  visual information fidelity, Information fusion 14~(2) (2013) 127--135.

\bibitem{jiaLLVIPVisibleinfraredPaired2023}
X.~Jia, C.~Zhu, M.~Li, W.~Tang, S.~Liu, W.~Zhou, {{LLVIP}}: {{A
  Visible-infrared Paired Dataset}} for {{Low-light Vision}} (Jun. 2023).
\newblock \href {http://arxiv.org/abs/2108.10831} {\path{arXiv:2108.10831}},
  \href {https://doi.org/10.48550/arXiv.2108.10831}
  {\path{doi:10.48550/arXiv.2108.10831}}.

\end{thebibliography}


%
%
%
\end{document}